\documentclass[doublecolumn,10pt] {IEEEtran}
\usepackage{cite}
\usepackage{amsmath,amssymb,amsfonts}
\usepackage{graphicx,color}
\usepackage{textcomp}
\usepackage{soul}
\sethlcolor{yellow}
\usepackage{xcolor} 
\definecolor{elsevierblue}{HTML}{2fb2e2} 

\usepackage[colorlinks=true,
            linkcolor=elsevierblue,  
            citecolor=elsevierblue,  
            urlcolor=elsevierblue    
           ]{hyperref}
\usepackage{enumitem}
\usepackage{multirow}
\usepackage[T1]{fontenc}
\usepackage{titlesec}
\usepackage{academicons}
\usepackage{xcolor}

\usepackage{subcaption} 
\usepackage{placeins}
\usepackage{booktabs}
\usepackage{tikz}
\usetikzlibrary{arrows.meta, positioning}
\usetikzlibrary{mindmap, shadows}
\usepackage{tabularx}
\usepackage{dblfloatfix} 

\usepackage[linesnumbered,ruled,vlined]{algorithm2e}

\definecolor{orcidlogocol}{HTML}{A6CE39}

\def\BibTeX{{\rm B\kern-.05em{\sc i\kern-.025em b}\kern-.08em
    T\kern-.1667em\lower.7ex\hbox{E}\kern-.125emX}}

\makeatletter
\renewenvironment{abstract}{%
  \begin{center}%
    {\normalfont\bfseries Abstract}%
  \end{center}%
  \normalfont 
}{%
  \par\vspace{0.5em}%
}

\let\oldIEEEkeywords\IEEEkeywords
\renewenvironment{IEEEkeywords}{%
  \oldIEEEkeywords
  \normalfont 
}{%
  \endlist
}
\makeatother

\begin{document}

\title{Green Energy Management for Sustainable Data Centers Using Deep Reinforcement Learning}

\author{%
\textbf{Abderaouf Bahi\textsuperscript{1\S}},
\textbf{Amel Ourici\textsuperscript{2}},
\textbf{Hasan Dinçer\textsuperscript{3,4,5}},
\textbf{Serhat Yüksel\textsuperscript{3,6}} and
\textbf{Akila Djebbar\textsuperscript{7}} \\[0.5em]
\textsuperscript{1}\small
Computer Science and Applied Mathematics Laboratory (LIMA),
Faculty of Science and Technology, Chadli Bendjedid University,
El Tarf 36000, Algeria \\[0.5em]
\textsuperscript{2}\small
Mathematical Modeling and Numerical Simulation Laboratory (LAM2SIN),
Faculty of Technology, Badji Mokhtar University,
Annaba 23000, Algeria \\[0.5em]
\textsuperscript{3}\small
School of Business, Istanbul Medipol University, Istanbul, Turkey \\[0.5em]
\textsuperscript{4}\small
Department of Economics and Management, Khazar University, Baku, Azerbaijan \\[0.5em]
\textsuperscript{5}\small
University College, Korea University, Seoul, Republic of Korea \\[0.5em]
\textsuperscript{6}\small
Clinic of Economics, Azerbaijan State University of Economics (UNEC), Baku, Azerbaijan \\[0.5em]
\textsuperscript{7}\small
Department of Computer Science, Faculty of Technology, LRI Laboratory,
Badji Mokhtar University, Annaba 23000, Algeria \\[0.5em]
\textsuperscript{\S}Corresponding author: \textbf{Abderaouf Bahi} (\href{mailto:a.bahi@univ-eltarf.dz}{a.bahi@univ-eltarf.dz})
}

\maketitle

\begin{abstract}
The exponential growth of digital services has positioned data centers among the most
energy-intensive infrastructures in the modern economy, raising critical concerns
regarding operational costs, carbon emissions, and the sustainable integration of
renewable energy sources. This paper proposes a novel Deep Reinforcement Learning
(DRL)-based energy management framework for data centers, designed to
dynamically coordinate solar photovoltaic generation, wind power, battery storage
systems, and conventional grid electricity under highly stochastic operational
conditions. The proposed framework formulates the energy management problem as a Markov
Decision Process and employs a Proximal Policy Optimization (PPO) agent augmented with
a hybrid Long Short-Term Memory and temporal attention architecture, enabling accurate
modeling of workload dynamics and renewable generation variability. A multi-objective
reward function jointly minimizes energy costs, carbon emissions, and service-level
agreement (SLA) violations while promoting efficient storage utilization. Extensive
experiments conducted on three datasets demonstrate that the proposed
framework achieves a 38\% reduction in energy costs compared to rule-based heuristics
and outperforms the strongest DRL baseline by 4.6\%, while maintaining an SLA violation
rate as low as 1.5\% and an energy efficiency of 83.7\%. Ablation studies confirm the
individual contribution of each architectural component, and hyperparameter sensitivity
analysis validates the robustness of the approach across a range of configurations.
\end{abstract}


\maketitle

\section{Introduction}
\label{sec:introduction}
The expansion of global digital services, and particularly the sustained growth of e-commerce platforms, has profoundly transformed the scale and operational demands of modern computing infrastructures. At the heart of this digital ecosystem lie large-scale data centers, which continuously process, store, and distribute massive volumes of information required to support online transactions, recommendation engines, logistics systems, and real-time customer interactions. As the digital economy expands, the energy requirements of these infrastructures have increased dramatically, positioning data centers among the most energy-intensive components of the modern technological landscape. Consequently, improving the energy efficiency of these facilities has become a critical scientific, economic, and environmental challenge. Continuous operation, high-performance computing requirements, and strict reliability constraints contribute to substantial electricity consumption and significant operational costs, while simultaneously generating a considerable carbon footprint \cite{1, 2, 3, 4}. 

In response to mounting environmental concerns and the global transition toward sustainable energy systems, considerable attention has been devoted to integrating renewable energy sources into the power infrastructure of data centers. Technologies such as solar photovoltaics and wind power offer promising pathways toward reducing greenhouse gas emissions and mitigating the environmental impact of large-scale computing infrastructures \cite{5, 6, 7, 8, 9}. These renewable resources provide clean and sustainable energy, yet their inherent intermittency introduces important operational challenges. Solar generation depends on weather conditions and diurnal cycles, while wind power is characterized by highly variable production patterns. Such fluctuations complicate the task of maintaining a stable and reliable power supply for data centers, which require uninterrupted energy availability to ensure continuous service delivery. Addressing these challenges requires the development of advanced energy management strategies capable of coordinating heterogeneous power sources, improving energy utilization, and minimizing reliance on conventional fossil-based electricity generation \cite{10, 11}.

Recent progress in artificial intelligence has opened new perspectives for addressing complex energy management problems in large-scale infrastructures. In particular, Deep Reinforcement Learning (DRL) has emerged as a powerful framework for sequential decision-making in dynamic and uncertain environments \cite{12, 13, 14}. Unlike traditional optimization approaches that rely on static models or predefined control rules, DRL enables adaptive learning through continuous interaction with the environment. This capability is particularly relevant for energy systems characterized by stochastic production patterns, fluctuating demand, and multiple interacting components. By leveraging large volumes of operational data including energy production levels, consumption patterns, and storage dynamics, DRL-based systems can learn effective policies for real-time power allocation and resource management. Such approaches enable intelligent decisions regarding when to prioritize renewable energy usage, store surplus generation, activate backup generators, or draw electricity from the main power grid \cite{15, 16, 17}. Through this adaptive control mechanism, DRL can significantly enhance operational efficiency while maximizing the utilization of renewable energy sources within data center infrastructures\cite{18,19,20,21,22,23,24,25}.

Renewable energy technologies themselves have undergone significant advancements over the past decade, resulting in improved efficiency, declining installation costs, and increased deployment worldwide. Solar photovoltaic systems convert solar radiation into electrical energy, while wind turbines transform kinetic wind energy into usable electricity \cite{26}. In practical deployments, these generation systems are often coupled with energy storage technologies, most notably battery systems, which allow excess energy produced during high-generation periods to be stored and later deployed during intervals of reduced renewable output \cite{27,28,29,30}. Despite these technological advances, managing hybrid renewable systems remains a complex task. Variability in generation, storage limitations, and unpredictable demand patterns require sophisticated coordination mechanisms capable of dynamically balancing supply and consumption. Traditional rule-based control strategies often struggle to cope with this level of complexity. Intelligent control mechanisms capable of learning optimal strategies directly from operational data therefore represent a promising direction for future energy management solutions \cite{31,32}. In this context, DRL-based control frameworks provide a particularly attractive approach for dynamically optimizing energy flows, facilitating seamless integration between renewable sources, storage systems, and conventional power infrastructures while improving overall system efficiency \cite{33,34,35,36}.

Motivated by these challenges and opportunities, this paper introduces a novel DRL-driven energy management framework specifically designed for data center environments. The proposed system aims to intelligently coordinate renewable energy sources, energy storage mechanisms, and conventional grid electricity in order to achieve improved energy efficiency while maintaining the strict reliability requirements of modern data centers. By continuously adapting its decision-making policy based on real-time system conditions, the proposed approach dynamically balances power flows across multiple energy sources. This enables the infrastructure to maximize renewable energy utilization when available, store surplus production when advantageous, and rely on conventional power only when necessary. Such adaptive energy management strategies have the potential to significantly reduce operational costs and environmental impact while enhancing the overall resilience and sustainability of data center operations. Nevertheless, several challenges remain, including the accurate forecasting of renewable generation patterns and the complexity associated with real-time optimization in highly dynamic energy environments \cite{10}.

The main contributions of this work can be summarized as follows:

\begin{itemize}
    \item We propose a novel DRL-based energy management framework for data centers that dynamically coordinates multiple energy sources including solar photovoltaic, wind power, battery storage systems, and grid electricity under stochastic conditions, while optimally balancing renewable usage, storage operations, and grid consumption to enhance energy efficiency and sustainability.
    \item We formulate the energy management problem as a Markov Decision Process (MDP), enabling adaptive and sequential decision-making that accounts for the uncertainty in both workload demand and renewable energy generation.
    \item We develop a Proximal Policy Optimization (PPO)-based agent enhanced with a hybrid Long Short-
    Term Memory (LSTM) and temporal attention architecture, allowing effective modeling of temporal dependencies and improving the prediction of energy demand and renewable supply fluctuations.
\end{itemize}

The remainder of this paper is organized as follows. Section 2 reviews the existing literature on renewable-powered data centers and intelligent energy management strategies. Section 3 presents the preliminaries. Section 4 details the proposed methodology, including its design principles and implementation. Section 5 describes the experimental setup and discusses the obtained results. Finally, Section 6 concludes the paper and outlines directions for future research.

\section{Related Work}
The growing integration of renewable energy sources into modern energy infrastructures has stimulated extensive research on intelligent control and optimization strategies. In particular, recent advances in artificial intelligence and data-driven methods have provided promising solutions for addressing the operational challenges introduced by the intermittent and stochastic nature of renewable energy systems. This section reviews the most relevant studies related to renewable energy management, forecasting, and data center optimization using DRL techniques.

\subsection{DRL for Renewable Energy Systems}
DRL has recently emerged as a powerful paradigm for decision-making and control in complex energy systems. Unlike traditional optimization approaches that rely on predefined models, DRL enables agents to learn optimal policies through interaction with dynamic environments. 

Li et al. \cite{li2023review} presented a comprehensive review of DRL techniques applied to modern renewable power systems. Their study highlighted how DRL can model power system operations as MDP and learn control policies capable of managing uncertainties introduced by renewable generation and flexible loads. The authors also discussed several DRL algorithms applied to operational control, emergency response, and system stability in modern power grids.

Similarly, Zhang et al. \cite{zhang2019energyconversion} proposed a DRL-based framework for optimizing energy conversion processes within integrated electrical and heating systems powered by renewable energy. Their approach formulated the energy conversion problem as a Markov decision process and employed the PPO algorithm to dynamically determine wind power conversion strategies. Experimental results demonstrated significant reductions in operating costs for system operators while maintaining system flexibility.

\subsection{Energy Forecasting and Renewable Generation Prediction}

Accurate forecasting of renewable energy generation plays a critical role in enabling efficient energy management and grid stability. Due to the variability of solar and wind resources, several studies have explored machine learning (ML) and deep learning (DL) techniques to improve prediction accuracy.

Benti et al. \cite{benti2023forecasting} reviewed recent advances in ML and DL methods for renewable energy forecasting. Their analysis highlighted the effectiveness of data-driven models in capturing complex nonlinear relationships in meteorological and generation data, significantly improving prediction accuracy compared with traditional statistical methods.

In a related study, Pradeep et al. \cite{pradeep2024deepfore} introduced DeepFore, a DRL-based forecasting framework for hybrid renewable energy systems. The proposed system integrates clustering techniques, feature optimization, and a Deep SARSA reinforcement learning (RL) algorithm to predict power generation. Their results showed that improved forecasting accuracy can significantly enhance the efficiency of renewable energy utilization.

\subsection{DRL for Energy Systems and Microgrid Optimization}

Beyond forecasting, DRL has also been applied to operational control and energy trading within distributed energy systems and microgrids. These environments involve complex interactions between renewable generation, storage systems, and dynamic electricity markets.

Harrold et al. \cite{harrold2022microgrid} explored the use of multi-agent DRL for managing renewable energy integration and energy trading within microgrids. Their approach used multiple agents to control hybrid energy storage systems while interacting with energy markets. The study demonstrated that multi-agent approaches achieved higher profitability and improved energy utilization compared with centralized control strategies.

Deng et al. \cite{deng2024dcrl} proposed a renewable building energy management method combining distributed Direct Current (DC) energy systems with DRL. Their framework employed the Soft Actor-Critic (SAC) algorithm to dynamically match energy demand with renewable supply while considering user satisfaction and operational flexibility. Experimental evaluations showed significant improvements in energy savings and renewable self-consumption.

Additionally, Jeong et al. \cite{jeong2023deepbid} introduced DeepBid, a DRL-based strategy for renewable energy bidding and battery control in real-time electricity markets. Their method simultaneously optimized bidding strategies and battery operations to maximize profits under uncertain generation and price conditions. The results demonstrated substantial improvements in revenue and reduction in deviation penalties.

\subsection{DRL for Sustainable Data Centers}

Data centers have become major energy consumers, motivating research on intelligent energy management strategies for sustainable computing infrastructures.

Zhang et al. \cite{zhang2022greendrl} proposed GreenDRL, a DRL-based framework designed to manage green data centers powered by renewable energy. Their system dynamically schedules workloads and energy consumption while accounting for environmental conditions and renewable generation variability. Experimental results demonstrated significant reductions in grid electricity usage compared with traditional scheduling approaches.

Jayanetti et al. \cite{jayanetti2024marl} introduced a multi-agent RL framework for renewable-energy-aware workflow scheduling across distributed cloud data centers. Their approach optimized task scheduling across geographically distributed infrastructures powered by both renewable and conventional energy sources. The proposed method achieved substantial reductions in workflow energy consumption while maintaining competitive execution times.

Zhao et al. \cite{zhao2025holistic} developed a RL-based framework for holistic energy optimization in sustainable cloud data centers. Their model integrated workload management and cooling optimization using a hybrid action space RL algorithm combined with Bayesian optimization. Experimental results showed notable reductions in both total energy consumption and brown energy usage.

More recently, Xiao and You \cite{xiao2026sustainable} investigated the integration of DRL-based control with renewable energy strategies for sustainable AI data centers. Their framework combined intelligent cooling control, demand response, and renewable energy utilization to simultaneously reduce operational costs, water consumption, and carbon emissions. The study demonstrated the potential of DRL-driven energy management for achieving scalable decarbonization of large AI infrastructures.

Overall, the literature demonstrates that DRL provides powerful capabilities for managing complex renewable-powered systems. However, despite these advances, several challenges remain, particularly regarding the coordinated management of renewable generation, energy storage, and real-time operational demands in large-scale data center environments. Addressing these challenges motivates the development of the proposed framework presented in this work.
Table~\ref{tab:ml_energy_comparison} summarizes representative ML and DL approaches applied to energy systems, smart grids, and renewable energy forecasting.  This comparative analysis is inspired by the format proposed by Anis et al. \cite{anis2026smartgridxai}. 

\begin{table*}[ht]
\centering
\caption{Overview of ML and DL approaches applied to energy systems and smart grid management \cite{anis2026smartgridxai}.}
\label{tab:ml_energy_comparison}
\scriptsize
\begin{tabular}{p{0.8cm} p{2.5cm} p{2.5cm} p{6cm} p{3.5cm}}
\hline
Ref. & Model & Application Domain & Key Outcome  & Limitations \\
\hline

\cite{yu2010decisiontree} & Decision Tree & Building energy demand prediction & Achieved $\sim$93\% training accuracy and 92\% testing accuracy in energy demand classification &  Limited to a single decision-tree architecture \\

\cite{chou2014modeling} & SVR, ANN, CART, CHAID, GLR & Heating and cooling load prediction & Hybrid models improved prediction accuracy; SVR and ANN performed best for cooling loads  & Increased complexity due to hybrid modeling strategies \\

\cite{sha2019hvac} & MLR, SVR, ANN & HVAC energy prediction & Reduced feature space while maintaining competitive prediction performance  & Lower accuracy for heating load prediction \\

\cite{peng2018occupancy} & Supervised and Unsupervised ML & Occupancy-aware HVAC control & Automated cooling setpoints based on occupancy prediction reduced energy consumption  & Performance depends heavily on accurate occupant data \\

\cite{fan2017cooling} & XGBoost + Deep Autoencoders & Cooling load forecasting & Deep feature extraction significantly improved short-term load prediction accuracy  & Higher computational requirements \\

\cite{badr2022federated} & Federated Learning + Hybrid DL & Energy forecasting with privacy preservation & Accurate energy prediction while protecting sensitive user data  & Implementation complexity due to federated learning architecture \\

\cite{ibrahem2020electricity} & ML + Functional Encryption & Electricity theft detection & Accurate theft detection and billing through encrypted smart meter data  & High complexity in encryption implementation \\

\cite{badr2023privacy} & Federated Learning + IPFE & Smart grid energy prediction & Accurate consumption and generation prediction with reduced communication overhead  & Complex deployment in real smart grid infrastructures \\

\cite{mocanu2016unsupervised} & RL + DBN & Short-term building energy prediction & Achieved $\sim$91\% prediction accuracy using automated feature extraction  & High computational cost for deep belief networks \\

\cite{sala2018activity} & RNN + ANFIS & HVAC thermal energy demand forecasting & Accurate short-term demand forecasting for HVAC systems  & Increased complexity of hybrid architecture \\

\cite{franovic2023decentralized} & MLP, XGBoost, SVM, GP & Smart grid stability prediction & ML models successfully predicted decentralized grid stability  & Limited analysis of interpretability and robustness \\

\cite{mancuso2021hierarchical} & Deep Neural Network & Hierarchical energy time-series forecasting & Improved accuracy for multi-level energy forecasting tasks  & Complex training due to hierarchical structure \\

\cite{shamshirband2019survey} & DL Survey & Renewable energy forecasting & Comprehensive overview of DL methods applied to solar and wind forecasting  & Does not propose a specific predictive model \\

\cite{cheng2018ensemble} & Ensemble RNN & Wind speed and power forecasting & High accuracy in 1-hour-ahead probabilistic wind forecasting  & Increased computational cost\\

\cite{liu2018wind} & EWT + LSTM + Elman NN & Wind speed forecasting & Multi-stage deep learning improved forecasting precision  & Higher computational complexity \\

\cite{kaloop2025metaheuristic} & ANN + PSO/ACO/ICA & Residential building energy estimation & High accuracy in early-stage building energy estimation  & Computationally intensive optimization process \\

\cite{henriksen2022load} & XGBoost + XAI & Residential load forecasting & Accurate load prediction with interpretable feature contributions  & Limited dataset scale \\

\cite{cifci2025interpretable} & Multiple ML Models + XAI & Smart grid load prediction & Demonstrated interpretable forecasting in decentralized grids & Limited benchmarking on real large-scale grids \\

\cite{sarker2024attention} & Attention-based DL + Federated Learning & Smart grid demand forecasting & Privacy-preserving load forecasting with strong prediction performance  & Increased model complexity and training cost \\

\hline
\end{tabular}
\end{table*}

\section{Preliminaries}
\paragraph{\normalfont\textbf{Definition 1}} (Data Center Energy Nodes)
Let the set of energy nodes in a data center be 
$N = \{n_1, n_2, \ldots, n_{|N|}\}$, 
where each node $n_i$ corresponds to a component such as a server cluster, cooling system, or energy storage device. 
Each node $n_i$ has a power demand or supply $p_t^{n_i}$ at time step $t$, which can originate from renewable sources, the electrical grid, or stored energy.

\paragraph{\normalfont\textbf{Definition 2}} (Workload and Energy Time Series)
Let the time horizon be $\mathcal{T} = \{1,2,\ldots,T\}$.  
For a given node $n_i$, the workload or energy consumption is represented as a time series:
$W^{n_i} = (w_1^{n_i}, w_2^{n_i}, \ldots, w_T^{n_i})$,  
where $w_t^{n_i}$ denotes the computational workload or energy usage at time $t$.  
Similarly, renewable energy generation at node $n_i$ is represented as $R^{n_i} = (r_1^{n_i}, r_2^{n_i}, \ldots, r_T^{n_i})$.

\paragraph{\normalfont\textbf{Definition 3}} (State Representation)
At each time step $t$, the environment state of the data center is defined as a vector
$\mathbf{s}_t = [p_t^{N}, W_t^{N}, R_t^{N}, S_t^{N}]$,  
where $p_t^{N}$ is the vector of power demands, $W_t^{N}$ is the workload vector, $R_t^{N}$ the renewable supply, and $S_t^{N}$ the stored energy levels.  
This state captures the real-time status of the data center’s energy and computational resources.

\paragraph{\normalfont\textbf{Definition 4}} (Action Space)
Let $\mathcal{A}$ denote the set of feasible actions for energy management.  
An action $a_t \in \mathcal{A}$ at time $t$ represents a control decision such as:  
- allocating workloads to servers,  
- charging/discharging energy storage,  
- drawing energy from the grid,  
- adjusting cooling or other operational parameters.

\paragraph{\normalfont\textbf{Definition 5}} (Reward Function)
The RL agent receives a reward $r_t$ at time $t$, designed to optimize energy efficiency and cost, as presented in Equation~\ref{eq:def5_reward}.
\begin{equation}
r_t = -\alpha \cdot C_t - \beta \cdot \text{SLA}_t
\label{eq:def5_reward}
\end{equation}
where $C_t$ is the energy cost at time $t$, $\text{SLA}_t$ is the service-level agreement violation at time $t$, and $\alpha, \beta$ are weighting factors balancing cost and quality of service.

\paragraph{\normalfont\textbf{Definition 6}} (Policy and DRL Agent)
Let $\pi_\theta(a_t|\mathbf{s}_t)$ denote a policy parameterized by $\theta$, which maps state $\mathbf{s}_t$ to action probabilities, as presented in Equation~\ref{eq:def6_policy}.
\begin{equation}
\pi_\theta(a_t|\mathbf{s}_t)
\label{eq:def6_policy}
\end{equation}
where $\pi_\theta$ is the stochastic policy, $a_t$ is the selected action at time $t$, $\mathbf{s}_t$ is the system state, and $\theta$ represents the learnable parameters of the policy.

The DRL agent aims to learn an optimal policy $\pi^*_\theta$ that maximizes the expected cumulative reward over the time horizon, as presented in Equation~\ref{eq:def6_objective}.
\begin{equation}
\pi^*_\theta = \arg\max_\theta \mathbb{E}\left[\sum_{t=1}^{T} r_t \right]
\label{eq:def6_objective}
\end{equation}
where $\pi^*_\theta$ denotes the optimal policy, $\mathbb{E}[\cdot]$ is the expectation operator, $r_t$ is the reward at time $t$, and $T$ is the time horizon.

\paragraph{\normalfont\textbf{Problem}} (Energy Management)
The objective is to determine the sequence of actions $\{a_1, a_2, \ldots, a_T\}$ that jointly minimize energy costs, reduce carbon emissions, maintain SLA compliance, and efficiently integrate renewable sources and storage systems in the data center.
\begin{figure*}[t]
    \centering
    \includegraphics[width=1\linewidth]{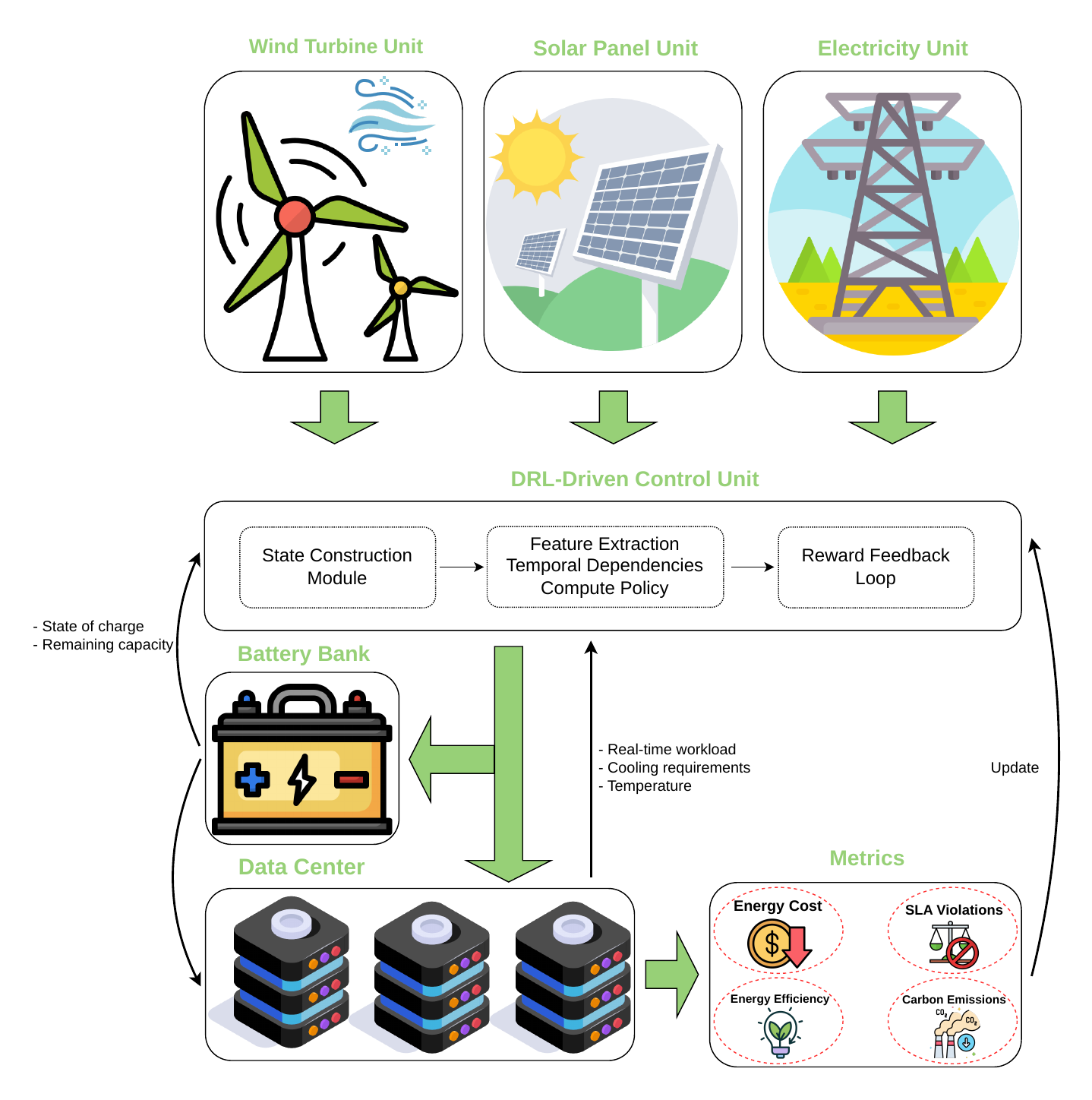}
    \caption{Overview of the proposed system.}
    \label{fig:architecture}
\end{figure*}

\section{Proposed Method}

This section introduces a comprehensive methodology for green energy management in e-commerce data centers based on DRL. In contrast to conventional approaches, the proposed framework is specifically designed to address three critical challenges inherent to modern data center operations: the intermittency of renewable energy sources, the temporal dynamics of computational workloads, and the complex trade-off between operational cost, carbon emissions, and SLA compliance. To this end, we formulate the problem as a MDP and design a customized reward function that enables the learning agent to achieve a balanced and sustainable energy management policy.

\subsection{System Overview}

The proposed system is structured around an integrated architecture that combines computational infrastructure, electriciy unit, renewable energy generation, energy storage, and an intelligent control mechanism. At its core, the data center hosts a variety of computational workloads that must be processed continuously, leading to dynamic energy consumption patterns influenced by user demand, processing intensity, and cooling requirements.

To formalize this behavior, the total energy consumption of the data center is expressed, as presented in Equation~\ref{eq:total_energy}, as the sum of computational and cooling demands.

\begin{equation}
P_{\text{total}}(t) = P_{\text{comp}}(t) + P_{\text{cool}}(t)
\label{eq:total_energy}
\end{equation}

where $P_{\text{total}}(t)$ denotes the total power consumption at time $t$, $P_{\text{comp}}(t)$ represents the computational energy demand, and $P_{\text{cool}}(t)$ corresponds to the cooling power required to maintain safe operating conditions.

The computational energy is directly driven by workload intensity. This relationship is modeled, as presented in Equation~\ref{eq:workload_energy}.

\begin{equation}
P_{\text{comp}}(t) = \kappa \cdot L(t)
\label{eq:workload_energy}
\end{equation}

where $\kappa$ is a proportionality constant reflecting system efficiency and $L(t)$ is the workload demand.

To reduce reliance on conventional energy sources, renewable energy systems such as solar panels and wind turbines are incorporated into the infrastructure. The total renewable energy generation is modeled, as presented in Equation~\ref{eq:renewable_total}.

\begin{equation}
E_{\text{renew}}(t) = E_{\text{solar}}(t) + E_{\text{wind}}(t)
\label{eq:renewable_total}
\end{equation}

where $E_{\text{solar}}(t)$ and $E_{\text{wind}}(t)$ denote the solar and wind energy generation at time $t$.

To mitigate this variability, energy storage systems, such as battery units, are deployed to store excess renewable energy during periods of high generation and redistribute it when production is insufficient. The storage dynamics are described, as presented in Equation~\ref{eq:soc_dyn}.

\begin{equation}
SOC(t+1) = SOC(t) + \eta_c P_{\text{charge}}(t) - \frac{1}{\eta_d} P_{\text{discharge}}(t)
\label{eq:soc_dyn}
\end{equation}

where $SOC(t)$ is the state-of-charge of the storage system, $\eta_c$ and $\eta_d$ are the charging and discharging efficiencies, and $P_{\text{charge}}(t)$ and $P_{\text{discharge}}(t)$ represent charging and discharging power.

The entire system is orchestrated by a centralized DRL-based control unit, which continuously observes the system state and determines optimal energy allocation strategies in real time. This control unit learns adaptive policies that dynamically balance multiple objectives, including efficiency, sustainability, and reliability. The overall architecture of the system is illustrated in Figure~\ref{fig:architecture}.

\subsection{MDP Formulation}

To formally model the decision-making process, the energy management problem is represented as a MDP defined by the tuple $(\mathcal{S}, \mathcal{A}, P, R, \gamma)$. At each time step $t$, the system is characterized by a state vector that captures both operational and environmental conditions. The state is defined as follows, as presented in Equation~\ref{eq:state}:

\begin{equation}
s_t = [L(t), E_{\text{renew}}(t), SOC(t), \theta_t]
\label{eq:state}
\end{equation}

where $L(t)$ represents the workload demand, $E_{\text{renew}}(t)$ denotes the predicted renewable energy generation, $SOC(t)$ corresponds to the storage level, and $\theta_t$ includes contextual features such as temperature and SLA indicators.

Based on the observed state, the agent selects an action that determines how energy resources are allocated across the system. The action vector is defined, as presented in Equation~\ref{eq:action}.

\begin{equation}
a_t = [P_{\text{grid}}(t), P_{\text{storage}}(t), \delta L(t)]
\label{eq:action}
\end{equation}

where $P_{\text{grid}}(t)$ is the power drawn from the grid, $P_{\text{storage}}(t)$ is the power supplied by storage, and $\delta L(t)$ represents workload shifting decisions.

To ensure system feasibility, the energy balance constraint must be satisfied at all times. This is formulated, as presented in Equation~\ref{eq:balance}.

\begin{equation}
P_{\text{total}}(t) = P_{\text{grid}}(t) + P_{\text{storage}}(t) + E_{\text{renew}}(t)
\label{eq:balance}
\end{equation}

where each term represents a component contributing to meeting the total demand.

The system dynamics are governed by the transition function $P(s_{t+1} \mid s_t, a_t)$, which captures the evolution of workloads, renewable energy availability, and storage behavior over time. These dynamics are inherently stochastic due to uncertainties in both demand and renewable generation.

To guide the learning process, a carefully designed reward function is introduced to reflect multiple, and often conflicting, objectives. The cost of grid energy is modeled, as presented in Equation~\ref{eq:cost}.

\begin{equation}
C_{\text{grid}}(t) = \lambda(t) \cdot P_{\text{grid}}(t)
\label{eq:cost}
\end{equation}

where $\lambda(t)$ denotes the time-dependent electricity price.

Similarly, carbon emissions are modeled, as presented in Equation~\ref{eq:emissions}.

\begin{equation}
\text{Emissions}(t) = \mu \cdot P_{\text{grid}}(t)
\label{eq:emissions}
\end{equation}

where $\mu$ is the emission factor associated with grid electricity.

Service reliability is quantified through SLA violations, defined as presented in Equation~\ref{eq:sla}.

\begin{equation}
\text{SLA\_viol}(t) = \max(0, L(t) - L_{\text{served}}(t))
\label{eq:sla}
\end{equation}

where $L_{\text{served}}(t)$ represents the workload actually processed.

The reward at time $t$ is defined, as presented in Equation~\ref{eq:reward}.

\begin{equation}
\begin{split}
r_t = & - \alpha \cdot C_{\text{grid}}(t) - \beta \cdot \text{Emissions}(t) \\
      & - \gamma \cdot \text{SLA\_viol}(t) - \delta \cdot |\Delta SOC(t)|
\end{split}
\label{eq:reward}
\end{equation}

where $\alpha$, $\beta$, $\gamma$, and $\delta$ are weighting coefficients controlling the importance of each objective, and $\Delta SOC(t) = SOC(t+1) - SOC(t)$ represents storage variation.

Finally, the discount factor $\gamma$ determines the importance of future rewards in the decision-making process, enabling the agent to balance short-term gains with long-term sustainability. The cumulative return is expressed, as presented in Equation~\ref{eq:return}.

\begin{equation}
R_t = \sum_{k=0}^{\infty} \gamma^k r_{t+k}
\label{eq:return}
\end{equation}

where $\gamma \in [0,1]$ controls the contribution of future rewards.

\subsection{Temporal Modeling of Workloads and Renewable Sources}

Accurately modeling temporal dependencies is essential for capturing both workload patterns and renewable energy variability. To address this, the proposed framework incorporates a hybrid recurrent architecture that combines LSTM networks with a temporal attention mechanism.

The LSTM component is responsible for learning long-term dependencies in sequential data. Its internal dynamics are defined, as presented in Equation~\ref{eq:lstm}.

\begin{equation}
h_t = \text{LSTM}(x_t, h_{t-1})
\label{eq:lstm}
\end{equation}

where $x_t$ is the input sequence, $h_t$ is the hidden state, and $h_{t-1}$ is the previous hidden state.

To enhance this, a temporal attention mechanism is introduced. The attention weights are computed, as presented in Equation~\ref{eq:attention}.

\begin{equation}
\alpha_i = \frac{\exp(e_i)}{\sum_j \exp(e_j)}
\label{eq:attention}
\end{equation}

where $e_i$ represents the relevance score of past observation $i$, and $\alpha_i$ denotes its normalized importance weight.

This hybrid design significantly enhances the model's ability to adapt to complex temporal behaviors.

\subsection{DRL Framework and PPO Agent}

The decision-making process is implemented using a PPO algorithm. The PPO objective function is defined, as presented in Equation~\ref{eq:ppo}.

\begin{equation}
L^{\text{PPO}}(\theta) = \mathbb{E} \left[ \min \left( 
\rho_t(\theta) A_t,\ 
\text{clip}(\rho_t(\theta), 1-\epsilon, 1+\epsilon) A_t 
\right) \right]
\label{eq:ppo}
\end{equation}

where $\rho_t(\theta) = \frac{\pi_\theta(a_t|s_t)}{\pi_{\theta_{\text{old}}}(a_t|s_t)}$ is the probability ratio, $A_t$ is the advantage function, and $\epsilon$ is a clipping parameter.

\subsection{Data Preprocessing}

The effectiveness of the proposed model relies heavily on the quality and consistency of the input data. Therefore, a rigorous preprocessing pipeline is applied prior to training. This process begins with data cleaning, where missing values and anomalies in energy consumption, renewable generation, and storage measurements are identified and corrected.

Subsequently, all numerical features are normalized using min-max scaling, defined as presented in Equation~\ref{eq:normalization}.

\begin{equation}
x' = \frac{x - x_{\min}}{x_{\max} - x_{\min}}
\label{eq:normalization}
\end{equation}

where $x$ is the original value and $x'$ is the normalized value.

Finally, heterogeneous data sources are temporally aligned and integrated into a unified dataset, ensuring that the DRL agent receives coherent and synchronized input.

Algorithm~\ref{alg:preprocessing} summarizes this procedure.
\begin{algorithm}[t!]
\small
\caption{Data Preprocessing for DRL Model Training}
\label{alg:preprocessing}

\textbf{Step 1: Input:} Raw data from energy sources ($E_{\text{renew}}(t)$), energy consumption patterns ($P_{dc}(t)$), and storage levels (SOC)\\
\textbf{Step 2: Output:} Preprocessed dataset ready for DRL model training\\
\textbf{Step 3: Data Cleaning:}\\
\hspace*{1em}\textbf{3.1:} Remove erroneous and incomplete data points: $D_{\text{cleaned}} = D_{\text{raw}} \setminus D_{\text{erroneous}}$\\
\hspace*{1em}\textbf{3.2:} Correct inconsistencies and handle outliers\\
\textbf{Step 4: Data Transformation:}\\
\hspace*{1em}\textbf{4.1:} Normalize numerical values: $X_{\text{norm}} = \frac{X - X_{\min}}{X_{\max} - X_{\min}}$\\
\hspace*{1em}\textbf{4.2:} Encode categorical variables: $V_{\text{encoded}} = f(V_{\text{categorical}})$\\
\hspace*{1em}\textbf{4.3:} Aggregate data into meaningful time intervals: $D_{\text{agg}}(t) = \sum_{i=t}^{t+\Delta t} D(i)$\\
\textbf{Step 5: Data Integration:}\\
\hspace*{1em}\textbf{5.1:} Merge data streams: $D_{\text{integrated}}(t) = [E_{\text{renew}}(t), P_{dc}(t), SOC(t)]$\\
\hspace*{1em}\textbf{5.2:} Create unified dataset for DRL model input: $D_{\text{final}} = D_{\text{integrated}}$\\
\textbf{Step 6: Finalize Preprocessing:}\\
\hspace*{1em}\textbf{6.1:} Ensure data is cohesive and correctly formatted for model training\\
\hspace*{1em}\textbf{6.2:} Output preprocessed data ready for DRL model training

\end{algorithm}
\subsection{Training and Validation}

The PPO agent is trained over multiple episodes that simulate realistic data center operations, allowing the model to progressively learn optimal energy management strategies under dynamic conditions. At each time step, the agent interacts with the environment by sampling actions from a stochastic policy conditioned on the observed state. This interaction process is formally defined, as presented in Equation~\ref{eq:policy_sampling}.

\begin{equation}
a_t \sim \pi_{\theta}(a_t \mid s_t)
\label{eq:policy_sampling}
\end{equation}

where $\pi_{\theta}$ denotes the parameterized policy with parameters $\theta$, $s_t$ is the current state, and $a_t$ is the sampled action.

Following action execution, the agent receives a reward signal that reflects the quality of the decision in terms of energy efficiency, environmental impact, and service reliability. The reward formulation is consistent with Equation~\ref{eq:reward}, and is accumulated over time to evaluate long-term performance. The cumulative discounted reward is computed, as presented in Equation~\ref{eq:return_training}.

\begin{equation}
R_t = \sum_{k=0}^{T} \gamma^k r_{t+k}
\label{eq:return_training}
\end{equation}

where $r_{t+k}$ is the reward at future time step $t+k$, $\gamma$ is the discount factor, and $T$ represents the episode horizon.

To reduce variance and stabilize learning, the advantage function is estimated. This is formulated, as presented in Equation~\ref{eq:advantage}.

\begin{equation}
A_t = R_t - V_{\phi}(s_t)
\label{eq:advantage}
\end{equation}

where $V_{\phi}(s_t)$ is the value function approximated by a neural network with parameters $\phi$.

The policy and value networks are then updated using the PPO clipped objective, ensuring stable and monotonic policy improvement. The optimization objective is defined, as presented in Equation~\ref{eq:ppo_train}.

\begin{equation}
L^{\text{PPO}}(\theta) = \mathbb{E} \left[ \min \left( 
\rho_t(\theta) A_t,\ 
\text{clip}(\rho_t(\theta), 1-\epsilon, 1+\epsilon) A_t 
\right) \right]
\label{eq:ppo_train}
\end{equation}

where $\rho_t(\theta) = \frac{\pi_{\theta}(a_t \mid s_t)}{\pi_{\theta_{\text{old}}}(a_t \mid s_t)}$ is the probability ratio between new and old policies, $A_t$ is the advantage estimate, and $\epsilon$ is a clipping parameter controlling the update magnitude.

In parallel, the value function is optimized by minimizing the prediction error. This is expressed, as presented in Equation~\ref{eq:value_loss}.

\begin{equation}
L^{\text{value}}(\phi) = \mathbb{E} \left[ \left( V_{\phi}(s_t) - R_t \right)^2 \right]
\label{eq:value_loss}
\end{equation}

where $V_{\phi}(s_t)$ is the predicted value and $R_t$ is the empirical return.

The overall training objective combines both policy and value losses, as presented in Equation~\ref{eq:total_loss}.

\begin{equation}
L_{\text{total}} = L^{\text{PPO}}(\theta) + c_1 L^{\text{value}}(\phi)
\label{eq:total_loss}
\end{equation}

where $c_1$ is a coefficient that balances the contribution of the value loss.

Algorithm~\ref{alg:drl_training} details the complete training procedure.

\begin{algorithm}[t!]
\small
\caption{DRL Model Training for Energy Management}
\label{alg:drl_training}

\textbf{Step 1: Input:} Preprocessed energy data ($E_{\text{renew}}(t)$), demand profile ($P_{dc}(t)$), and storage constraints (SOC)\\
\textbf{Step 2: Output:} Trained PPO model for real-time energy management\\

\textbf{Step 3: Environment and Agent Initialization:}\\
\hspace*{1em}\textbf{3.1:} Initialize the smart grid environment using integrated energy data\\
\hspace*{1em}\textbf{3.2:} Define the state space $s_t = [E_{\text{renew}}(t), P_{dc}(t), SOC(t)]$\\
\hspace*{1em}\textbf{3.3:} Define the action space $a_t \in \mathcal{A}$ representing energy dispatch decisions\\
\hspace*{1em}\textbf{3.4:} Initialize the PPO agent with neural network architecture \\

\textbf{Step 4: Reward Function Design:}\\
\hspace*{1em}\textbf{4.1:} Define reward as a weighted combination of energy efficiency, operational cost, and stability\\
\hspace*{1em}\textbf{4.2:} $r_t = \alpha \cdot \eta_{\text{energy}} - \beta \cdot C_{\text{grid}} - \gamma \cdot \Delta SOC$\\

\textbf{Step 5: Training Process:}\\
\hspace*{1em}\textbf{5.1:} Initialize PPO policy $\pi_{\theta}(a_t|s_t)$ and value function $V_{\phi}(s_t)$\\
\hspace*{1em}\textbf{5.2:} \textbf{For each training episode:}\\
\hspace*{2em}\textbf{5.2.1:} Reset the environment and observe initial state $s_0$\\
\hspace*{2em}\textbf{5.2.2:} \textbf{While} the terminal condition is not reached:\\
\hspace*{3em}\textbf{(a)} Select action $a_t$ according to current policy $\pi_{\theta}(a_t|s_t)$\\
\hspace*{3em}\textbf{(b)} Execute action and observe next state $s_{t+1}$ and reward $r_t$\\
\hspace*{3em}\textbf{(c)} Store transition $(s_t, a_t, r_t, s_{t+1})$ in the replay buffer\\
\hspace*{2em}\textbf{5.2.3:} Update policy parameters using PPO clipped objective function\\

\textbf{Step 6: Policy Evaluation and Optimization:}\\
\hspace*{1em}\textbf{6.1:} Evaluate the trained policy on unseen operational scenarios\\
\hspace*{1em}\textbf{6.2:} Analyze performance metrics (energy cost reduction, efficiency, stability)\\
\hspace*{1em}\textbf{6.3:} Fine-tune hyperparameters for deployment readiness\\

\textbf{Step 7: Deployment:}\\
\hspace*{1em}\textbf{7.1:} Export the trained PPO model for real-time decision-making\\
\hspace*{1em}\textbf{7.2:} Integrate the model into the energy management system

\end{algorithm}

\section{Experiments}
\subsection{Experimental Setup}

\subsubsection{Datasets}

The proposed framework is evaluated using three datasets that cover complementary aspects of data center energy management: renewable energy generation, workload demand, and electricity pricing.

\textbf{(1) Solar Energy Generation Dataset (Kaggle).}\footnote{\url{https://www.kaggle.com/datasets/anikannal/solar-power-generation-data}} This dataset contains high-frequency solar power generation records collected from two solar plants in India over a 34-day period, including DC/AC power output and irradiation measurements sampled every 15 minutes. It is used to model photovoltaic generation dynamics in our simulation environment.

\textbf{(2) Wind Power Forecasting Dataset (Kaggle).}\footnote{\url{https://www.kaggle.com/datasets/theforcecoder/wind-power-forecasting}} This dataset provides 4.5 years of hourly wind turbine measurements including wind speed, wind direction, theoretical power curve values, and actual power output. It enables realistic modeling of wind energy variability and intermittency.

\textbf{(3) Grid Electricity Pricing and Demand Dataset.} 
This dataset was collected from publicly available reports and local measurements in Algeria, reflecting the characteristics of a conventional power grid supplying urban infrastructures. It includes hourly records of electricity prices, grid supply availability, and demand profiles over a one-year period. The dataset captures temporal variations such as peak/off-peak tariffs, seasonal demand fluctuations, and grid dependency patterns.

The three datasets were temporally aligned to a common 1-hour resolution and preprocessed using the pipeline described in Algorithm~\ref{alg:preprocessing}. The combined dataset spans approximately 8,760 time steps, corresponding to one full year of operation, split into 70\% for training, 15\% for validation, and 15\% for testing.

\subsubsection{Baselines}

To rigorously evaluate the proposed DRL-based energy management framework, we compare it against six representative baselines. All baselines are implemented within the same simulation environment using a unified codebase.

\textbf{(1) Rule-Based Heuristic (RBH).} A deterministic control strategy that prioritizes renewable energy consumption, charges storage when generation exceeds demand, and falls back to grid power otherwise. This baseline represents conventional operational practice and serves as a lower bound for intelligent approaches.

\textbf{(2) Q-Learning (QL)} \cite{watkins1992q}. A classical tabular RL method that discretizes the state and action spaces. The agent learns a Q-table through $\epsilon$-greedy exploration. This baseline demonstrates the limitations of non-parametric RL in high-dimensional energy environments.

\textbf{(3) Deep Q-Network (DQN)} \cite{mnih2015dqn}. A foundational DRL approach that approximates the Q-function using a neural network and employs experience replay and target networks for stability. DQN serves as a key comparison point for assessing the benefit of policy-gradient methods over value-based RL.

\textbf{(4) Asynchronous Actor-Critic (A2C)} \cite{mnih2016a3c}. An asynchronous policy-gradient method that simultaneously learns a policy network and a value function. A2C enables continuous action spaces and provides a direct comparison to our PPO-based formulation.

\textbf{(5) Soft Actor-Critic (SAC)} \cite{haarnoja2018sac}. An off-policy maximum-entropy RL algorithm that encourages exploration through entropy regularization. SAC has demonstrated strong performance in continuous control tasks and represents the current state of the art in model-free DRL for energy systems \cite{deng2024dcrl}.

\textbf{(6) GreenDRL} \cite{zhang2022greendrl}. A recent DRL-based framework specifically designed for renewable-powered data center scheduling. GreenDRL combines workload management with renewable-aware energy dispatching and constitutes the most directly comparable prior work to our proposed approach.

\begin{table*}[t!]
\centering
\caption{Architecture of the PPO model.}
\label{tab:ppo_architecture}

\begin{tabular}{p{3cm}p{2cm}p{2.5cm}p{2.5cm}p{5.5cm}}
\hline
Layer       & Type      & Dimension               & Activation Function & Hyperparameters \\ \hline
Input Layer           & -                    & -                                & -                           & -                        \\ 
Convolutional Layer 1 & Convolutional        & 64 filters, 3x3                  & ReLU                        & Stride: 1, Padding: Same \\ 
Convolutional Layer 2 & Convolutional        & 128 filters, 3x3                 & ReLU                        & Stride: 2, Padding: Same \\ 
Recurrent Layer       & LSTM                 & 256 units                        & Tanh                        & Dropout: 0.5, Return Sequences: True \\ 
Fully Connected 1     & Dense                & 512 units                        & ReLU                        & -                        \\ 
Fully Connected 2     & Dense                & 256 units                        & ReLU                        & -                        \\ 
Output Layer (Policy) & Dense (Policy)       & Action Space Size                & Softmax                     & -                        \\ 
Output Layer (Value)  & Dense (Value)        & 1                                & Linear                      & -                        \\ \hline
\end{tabular}
\end{table*}
\subsubsection{Evaluation Metrics}
The performance of the proposed system is evaluated using several key metrics that assess energy efficiency, cost savings, and environmental impact. These metrics also allow for a comparative analysis against prevailing benchmarks and prior research:

\textbf{Energy Cost:} Measures the total expenditure on energy consumption, accounting for both renewable and grid energy. The goal is to minimize energy costs by optimizing the use of renewable resources and storage systems. The metric is represented as Equation~\ref{eq:energycost}.
\begin{equation}
\label{eq:energycost}
    \text{Energy Cost} = \sum_{t=1}^{T} (E_{grid}(t) \times C_{grid} + E_{storage}(t) \times C_{storage})
\end{equation}
where $E_{grid}(t)$ is the energy drawn from the grid at time $t$, $C_{grid}$ is the cost per unit of grid energy, $E_{storage}(t)$ is the energy taken from storage, and $C_{storage}$ is the cost associated with using energy from storage.

\textbf{SLA Violations:} Refers to instances where the system fails to meet predefined energy supply requirements, potentially disrupting the data center’s operations. The DRL model aims to minimize these violations, represented by Equation~\ref{eq:slasla}.
\begin{equation}
\label{eq:slasla}
    \text{SLA Violations} = \sum_{t=1}^{T} I(E_{demand}(t) > E_{supplied}(t))
\end{equation}
where $E_{demand}(t)$ is the energy demand at time $t$, $E_{supplied}(t)$ is the energy supplied at time $t$, and $I(.)$ is an indicator function that equals 1 when an SLA violation occurs, and 0 otherwise.

\textbf{Energy Efficiency:} Measures how effectively energy resources are utilized to meet the data center’s demands while minimizing waste. The metric is represented as Equation~\ref{eq:enerener}.
\begin{equation}
\label{eq:enerener}
    \text{Energy Efficiency} = \frac{E_{renewable}(t)}{E_{total}(t)}
\end{equation}
where $E_{renewable}(t)$ is the energy supplied from renewable sources at time $t$, and $E_{total}(t)$ is the total energy consumption.

\textbf{Carbon Emissions:} This metric measures the environmental impact of energy consumption, primarily focusing on the carbon footprint from non-renewable energy use. It is calculated as Equation~\ref{eq:crbemi}.
\begin{equation}
\label{eq:crbemi}
    \text{Carbon Emissions} = \sum_{t=1}^{T} E_{grid}(t) \times F_{emission}
\end{equation}
where $E_{grid}(t)$ is the energy drawn from the grid at time $t$, and $F_{emission}$ is the carbon emission factor.

\textbf{Cumulative Reward:} In RL, the cumulative reward represents the total rewards accumulated by the DRL agent over the training or testing period, quantifying its overall performance. It is calculated as Equation~\ref{eq:cumur}.
\begin{equation}
\label{eq:cumur}
    \text{Cumulative Reward} = \sum_{t=1}^{T} R(t)
\end{equation}
where $R(t)$ is the reward received at time $t$, a function of energy cost, SLA violations, energy efficiency, and carbon emissions.

\textbf{Success Rate:} This metric measures the proportion of scenarios where the DRL model successfully meets predefined goals, such as achieving a specific energy efficiency level or maintaining SLA compliance. It is represented as Equation~\ref{eq:scrate}.
\begin{equation}
\label{eq:scrate}
    \text{Success Rate} = \frac{\text{Number of Successful Scenarios}}{\text{Total Number of Scenarios}}
\end{equation}

\subsubsection{Model Architecture}

The neural architecture is designed to effectively process high-dimensional and temporally correlated inputs. Initially, convolutional layers are employed to extract spatial and structural features from the input data. The transformation performed by a convolutional layer is defined, as presented in Equation~\ref{eq:conv}.

\begin{equation}
x^{(l+1)} = \sigma(W^{(l)} * x^{(l)} + b^{(l)})
\label{eq:conv}
\end{equation}

where $x^{(l)}$ is the input feature map at layer $l$, $W^{(l)}$ represents convolutional filters, $*$ denotes the convolution operation, $b^{(l)}$ is the bias term, and $\sigma$ is a nonlinear activation function.

To capture temporal dependencies, the extracted features are passed through hybrid LSTM-Attention layers. The LSTM hidden state evolution is expressed, as presented in Equation~\ref{eq:lstm_arch}.

\begin{equation}
h_t = \text{LSTM}(x_t, h_{t-1})
\label{eq:lstm_arch}
\end{equation}

where $x_t$ is the input at time $t$, $h_t$ is the hidden state, and $h_{t-1}$ is the previous hidden state.

The attention mechanism further refines this representation by assigning importance weights to temporal features. The attention output is computed, as presented in Equation~\ref{eq:attention_arch}.

\begin{equation}
z_t = \sum_{i=1}^{t} \alpha_i h_i
\label{eq:attention_arch}
\end{equation}

where $\alpha_i$ represents the attention weight associated with hidden state $h_i$, and $z_t$ is the context vector summarizing relevant temporal information.

Finally, fully connected layers are used to approximate both the policy and value functions. The mapping is defined, as presented in Equation~\ref{eq:fc}.

\begin{equation}
y = f(Wx + b)
\label{eq:fc}
\end{equation}

where $x$ is the input feature vector, $W$ and $b$ are learnable parameters, and $f$ is a nonlinear activation function.

Hyperparameters were tuned via systematic experimentation to maximize cumulative reward and energy efficiency. This optimization process can be formulated, as presented in Equation~\ref{eq:hyperparam}.

\begin{equation}
\theta^* = \arg\max_{\theta} \mathbb{E}[R_t]
\label{eq:hyperparam}
\end{equation}

where $\theta$ represents the set of hyperparameters and $R_t$ is the cumulative reward.

The final configuration is summarized in Table~\ref{tab:ppo_architecture}.

\subsection{Experimental Results}

Table~\ref{tab:results_comparison} presents the performance comparison of the proposed DRL framework against all baselines. Results are averaged over five independent runs with different random seeds.

\begin{table*}[t!]
\centering
\caption{Comparison of energy management performance across baselines and the proposed DRL framework.}
\label{tab:results_comparison}
\small
\begin{tabular}{lcccccc}
\hline
Method & Energy Cost (\$) & SLA Violations (\%) & Energy Efficiency (\%) & Carbon Emissions (kg) & Success Rate (\%) \\
\hline
RBH           & 1842.3 & 6.8  & 61.2 & 487.4 & 73.1 \\
Q-Learning    & 1623.7 & 5.1  & 67.4  & 421.8 & 78.4 \\
DQN           & 1421.5 & 3.9  & 71.8  & 378.6 & 82.7 \\
A2C           & 1356.2 & 3.2  & 74.3  & 354.2 & 84.9 \\
SAC           & 1274.8 & 2.6  & 77.1  & 331.7 & 87.3 \\
GreenDRL      & 1198.3 & 2.1  & 79.6  & 312.5 & 89.1 \\
\textbf{This work} & \textbf{1143.4} & \textbf{1.5} & \textbf{83.7}  & \textbf{291.8} & \textbf{92.4} \\
\hline
\end{tabular}
\end{table*}
\begin{figure*}[h!]
    \centering
    \includegraphics[width=\linewidth]{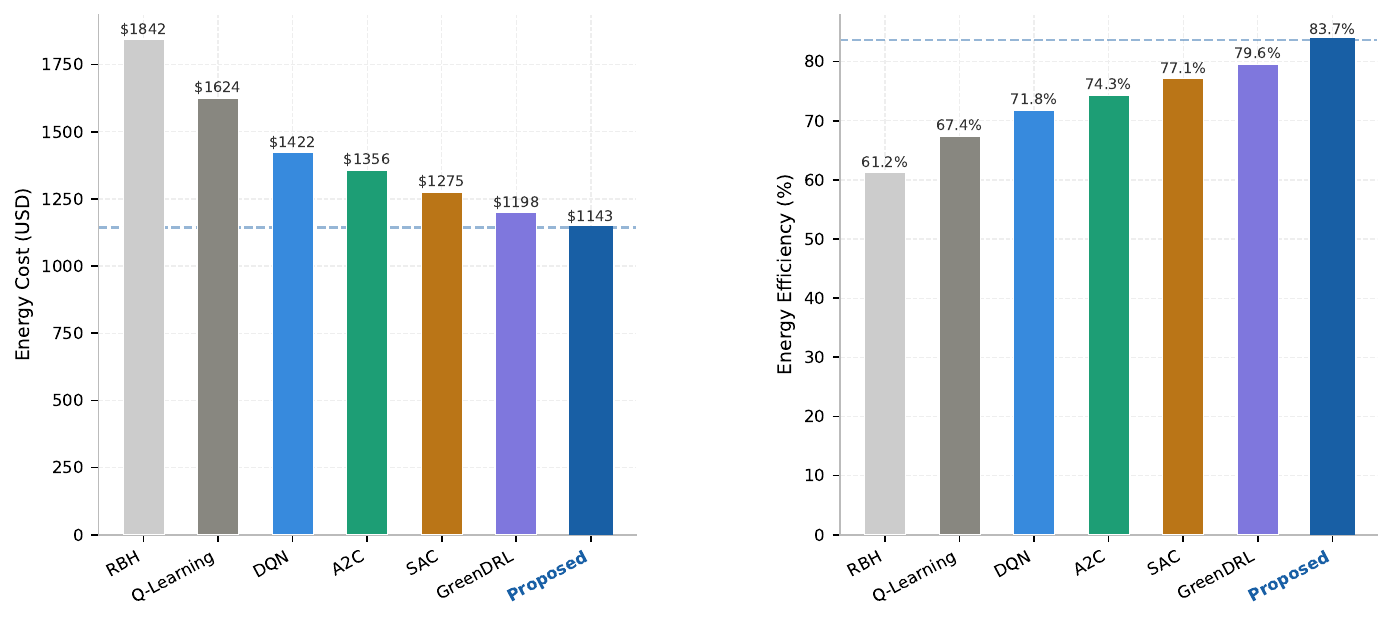}
    \caption{Comparison of energy cost (USD) and energy efficiency (\%) across all baseline
    methods and the proposed framework.}
    \label{fig:method_comparison}
\end{figure*}
The proposed framework consistently outperforms all baselines across every metric. Compared to the rule-based heuristic, the proposed approach reduces energy costs by approximately 38\%, decreases SLA violations from 6.8\% to 1.5\%, and cuts carbon emissions by 40.1\%. Against the strongest baseline, the proposed method achieves an additional 4.6\% cost reduction and a 6.6\% improvement in carbon efficiency, demonstrating the benefit of combining LSTM-based temporal modeling with the PPO policy optimization.

The gap between DQN and the policy-gradient methods confirms that continuous action spaces are better suited for fine-grained energy allocation. The entropy-based exploration of SAC partially compensates for the absence of temporal modeling, yet it still underperforms the proposed approach due to the lack of attention-based workload forecasting. Overall, these results validate the three design choices of the proposed framework: (i) the PPO learning objective for stable policy updates, (ii) the LSTM-Attention module for capturing temporal workload and generation patterns, and (iii) the multi-objective reward function balancing cost, reliability, and sustainability.

Figure~\ref{fig:method_comparison} presents a side-by-side comparison of energy cost and
energy efficiency across all evaluated methods. The proposed framework achieves the lowest
energy cost of \$1,143.4, representing a reduction of approximately 38\% compared to the
RBH and 4.6\% compared to GreenDRL. In
parallel, the proposed method attains the highest energy efficiency of 83.7\%, confirming
its ability to maximise the utilisation of renewable energy sources while minimising
reliance on grid electricity.

\begin{figure*}[t!]
    \centering
    \includegraphics[width=\linewidth]{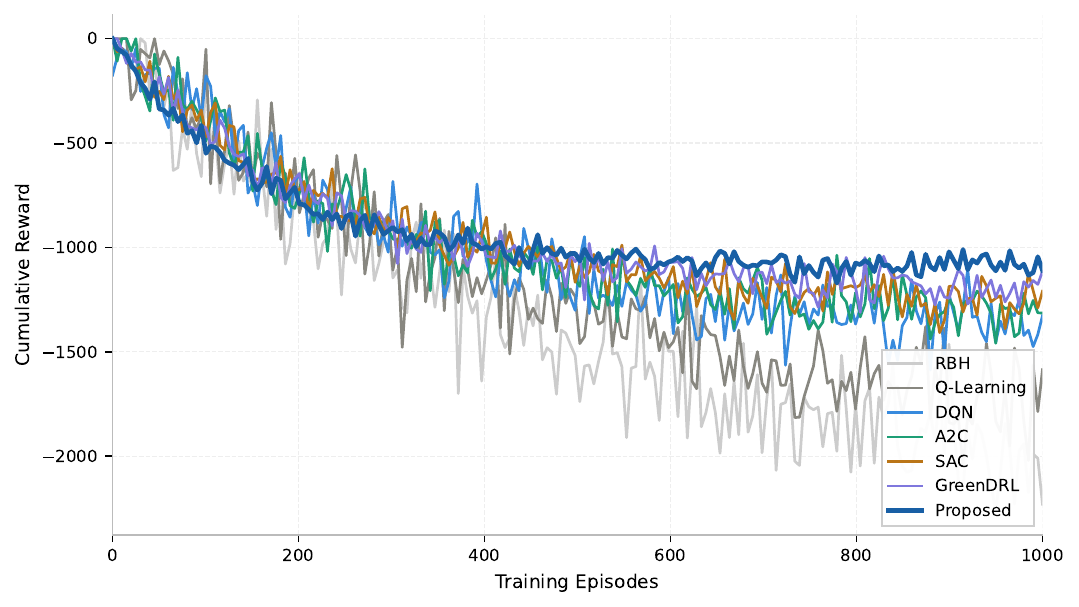}
    \caption{Cumulative reward per training episode for all methods.}
    \label{fig:learning_curves}
\end{figure*}

Figure~\ref{fig:learning_curves} illustrates the convergence behaviour of each DRL agent
over the course of training. The proposed PPO-based framework with LSTM-Attention
consistently converges faster and stabilises at a higher cumulative reward than all
competing methods. While DQN and A2C exhibit slower convergence and higher variance in
their reward trajectories, the proposed approach achieves stable improvement within fewer
episodes, demonstrating the benefit of the hybrid recurrent architecture and the clipped
PPO objective for stable policy updates in dynamic energy environments.
\begin{figure*}[t!]
    \centering
    \includegraphics[width=\linewidth]{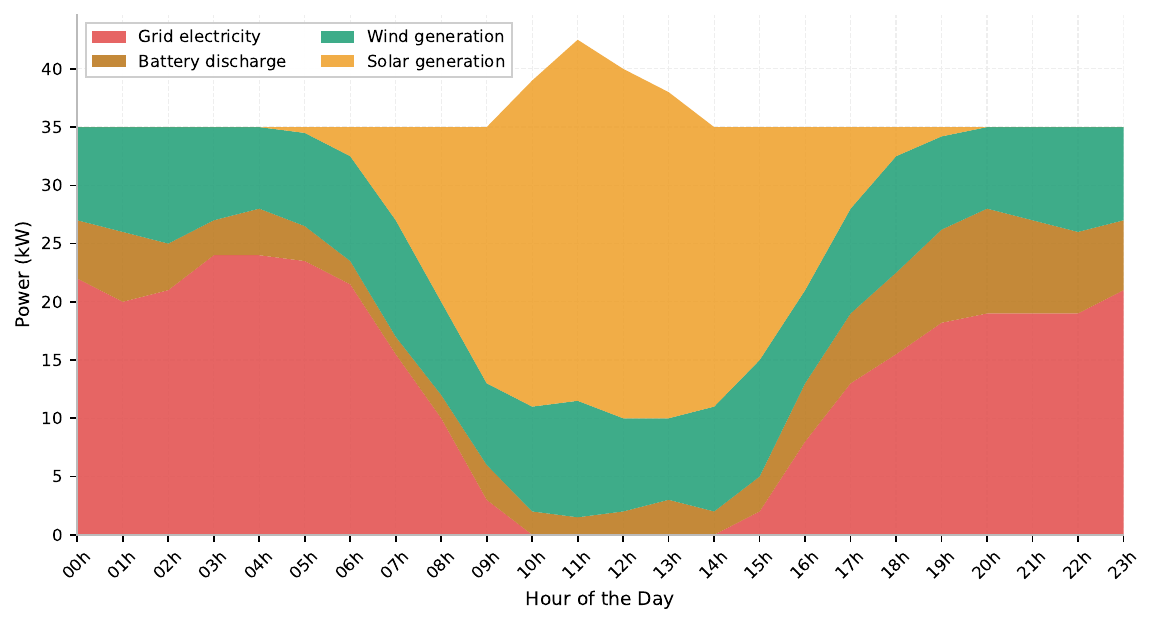}
    \caption{Temporal decomposition of energy sources managed by the proposed framework
    over a representative 24-hour period.}
    \label{fig:energy_mix}
\end{figure*}
Figure~\ref{fig:energy_mix} depicts the temporal distribution of energy sources across a
representative 24-hour operational period under the control of the proposed framework.
Solar generation dominates during daylight hours, while wind
power provides a relatively steady contribution throughout the day. Battery discharge is strategically deployed during transitional periods to smooth supply fluctuations, and grid
electricity is drawn only when renewable and storage resources are insufficient to meet
demand. This behaviour confirms that the DRL agent has learned to effectively prioritise
clean energy sources and limit grid dependence to off-peak or critical intervals.

\subsubsection{Ablation Study}

To isolate the contribution of each component of the proposed framework, we conduct an ablation study by systematically removing individual modules and evaluating the resulting performance degradation. Four ablated variants are considered:

\begin{itemize}
    \item \textbf{w/o LSTM}: Replaces the LSTM recurrent module with a standard feedforward network, removing temporal sequence modeling.
    \item \textbf{w/o Attention}: Removes the temporal attention mechanism while retaining the LSTM component.
    \item \textbf{w/o Emissions}: Excludes the carbon emissions penalty term from the reward function.
    \item \textbf{w/o Storage}: Disables the energy storage dispatch component, leaving only grid and renewable sources.
\end{itemize}
\begin{figure*}[h]
    \centering
    \includegraphics[width=\linewidth]{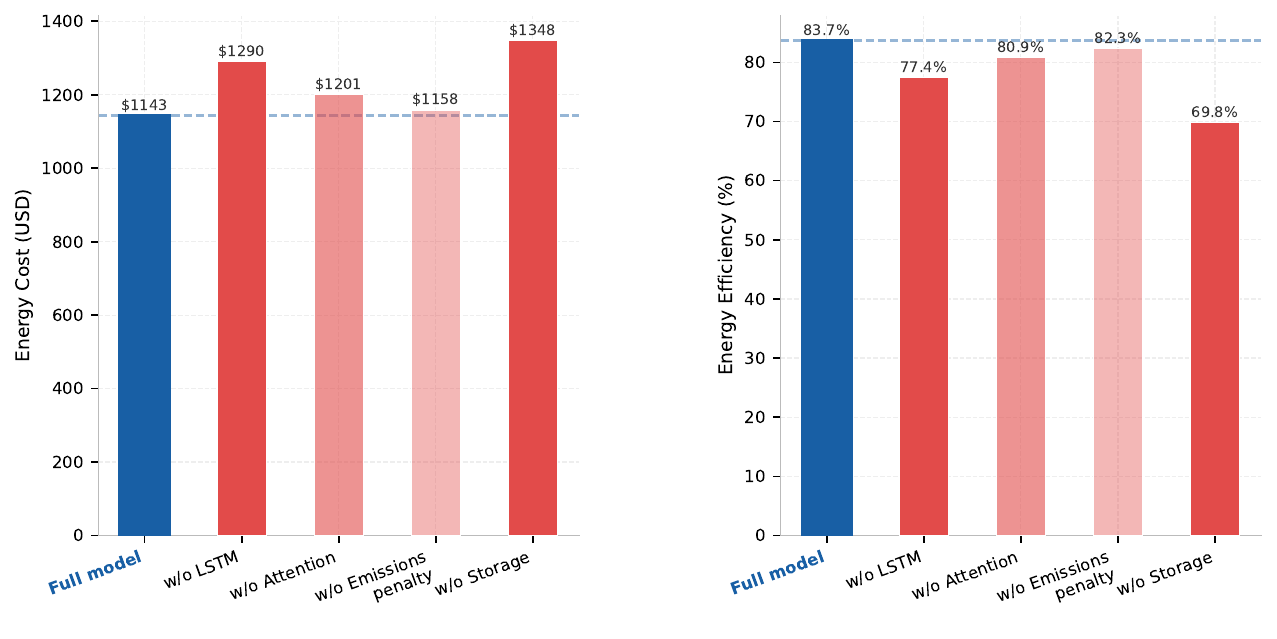}
    \caption{Ablation study results comparing the full model against four component-removed
    variants in terms of energy cost and energy efficiency.}
    \label{fig:ablation}
\end{figure*}
\begin{table}[t!]
\centering
\caption{Ablation study results.}
\label{tab:ablation}
\footnotesize
\begin{tabular}{p{1.8cm} p{1cm} p{1cm} p{1.3cm} p{1.4cm}}
\hline
Variant & Energy Cost & SLA Violations & Energy Efficency & Carbon Emissions\\
\hline
Full Model              & \textbf{1143.4} & \textbf{1.5\%} & \textbf{83.7\%} & \textbf{291.8} \\
w/o LSTM                & 1289.7 & 2.8\% & 77.4\% & 338.5 \\
w/o Attention           & 1201.3 & 2.1\% & 80.9\% & 311.2 \\
w/o Emissions           & 1158.2 & 1.6\% & 82.3\% & 367.4 \\
w/o Storage             & 1347.8 & 3.4\% & 69.8\% & 352.6 \\
\hline
\end{tabular}
\end{table}

Table~\ref{tab:ablation} confirms that all components contribute meaningfully to overall performance. Removing the LSTM module causes the largest degradation in energy cost (+12.8\%), highlighting the critical role of temporal sequence modeling for anticipating workload and generation peaks. Notably, the variant without the emissions penalty achieves slightly higher energy costs but at the expense of a 25.9\% increase in carbon emissions, demonstrating the trade-off governed by the weighting coefficient $\beta$. The removal of storage dispatching produces the worst overall configuration, underscoring the necessity of battery integration for buffering renewable variability.

Figure~\ref{fig:ablation} reports the results of the ablation study conducted to quantify
the individual contribution of each architectural component. Removing the LSTM module
induces the most significant performance degradation, with energy cost increasing by
12.8\% and efficiency dropping by 6.3 percentage points, confirming the critical role of
temporal sequence modelling in capturing workload and generation dynamics. The removal of
the attention mechanism results in a moderate degradation of 5.1\% in cost, while
disabling the carbon emission penalty slightly increases energy cost (~1.3\%) and substantially
increases the carbon footprint. Finally, the variant without storage dispatching exhibits
the worst efficiency (69.8\%), underscoring the necessity of battery integration for
buffering renewable intermittency.

\subsubsection{Hyperparameter Sensitivity Analysis}

To understand how key hyperparameters influence system performance, we conduct a sensitivity analysis by varying one parameter at a time while keeping the others fixed at their optimal values. Four critical hyperparameters are examined: the PPO clipping parameter $\epsilon$, the discount factor $\gamma$, the reward weighting coefficient $\alpha$ (energy cost), and the LSTM hidden size.

\textbf{Clipping parameter $\epsilon$}: Values in $\{0.1, 0.2, 0.3, 0.4\}$ are tested. Performance peaks at $\epsilon = 0.2$, which balances conservative policy updates with sufficient exploration. Larger values introduce instability, while smaller values slow convergence.

\textbf{Discount factor $\gamma$}: Values in $\{0.90, 0.95, 0.97, 0.99\}$ are evaluated. A value of $\gamma = 0.97$ yields the best trade-off between short-term cost minimization and long-term sustainability planning. Lower values lead to myopic policies that over-discharge storage, whereas very high values slow learning convergence.

\textbf{Cost weighting coefficient $\alpha$}: Values in $\{0.3, 0.5, 0.7, 1.0\}$ are explored. Setting $\alpha = 0.5$ provides the best balance between energy cost reduction and SLA compliance. Higher values aggressively minimize grid usage but increase violation rates due to insufficient safety margins.

\textbf{LSTM hidden size}: Sizes in $\{64, 128, 256, 512\}$ are tested. A hidden size of 256 units achieves the best performance, offering sufficient representational capacity to model complex temporal patterns without overfitting on the available training data.

These results confirm that the proposed framework is robust within reasonable hyperparameter ranges and that the chosen configuration is near-optimal.

\begin{figure*}[h]
    \centering
    \includegraphics[width=\linewidth]{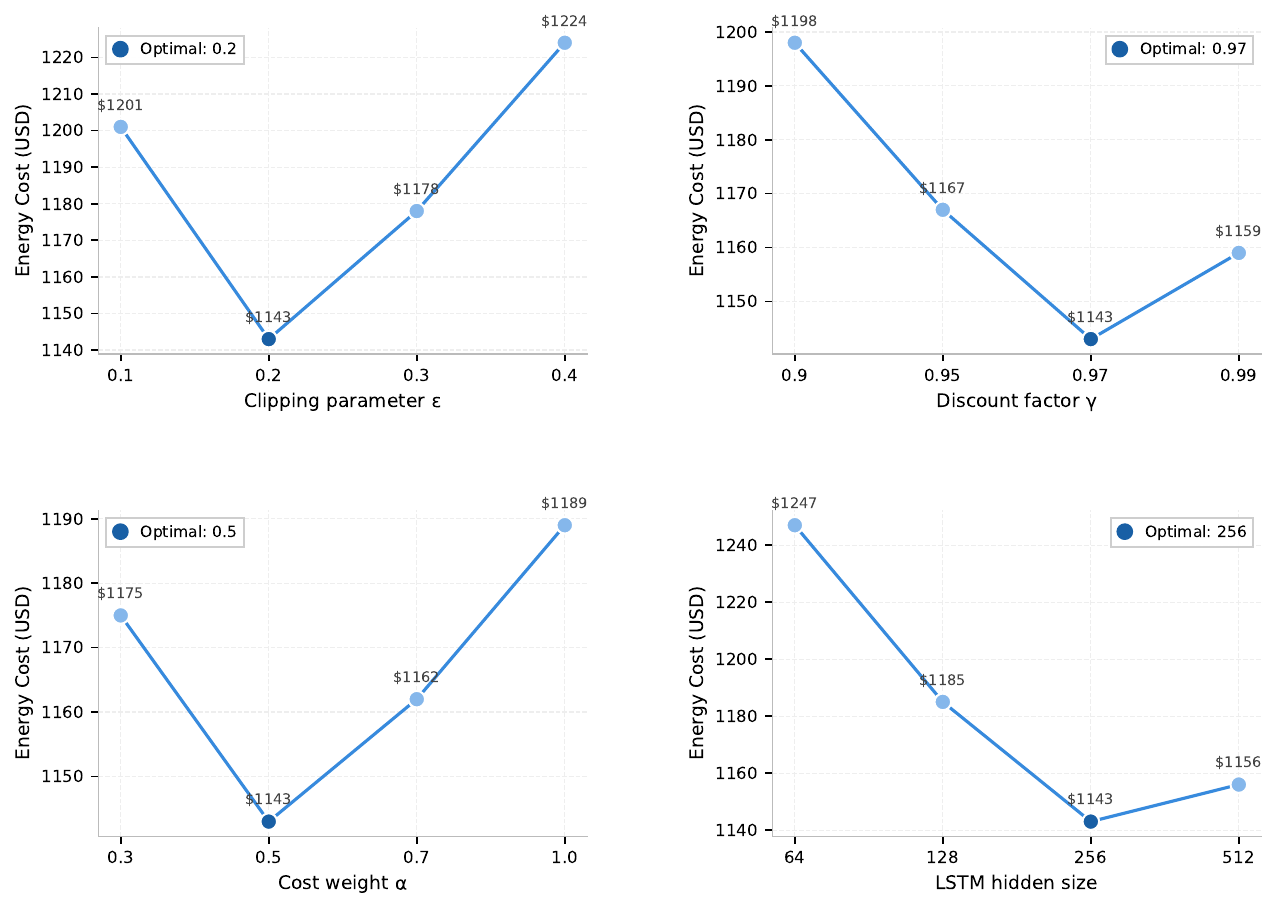}
    \caption{Sensitivity analysis of four key hyperparameters: the PPO clipping parameter
    $\epsilon$, the discount factor $\gamma$, the cost weighting coefficient $\alpha$, and
    the LSTM hidden size.}
    \label{fig:hyperparameter}
\end{figure*}

Figure~\ref{fig:hyperparameter} presents the sensitivity of the proposed framework to
four critical hyperparameters, each varied independently while holding the others fixed
at their optimal values. The performance curves exhibit clear optima at $\epsilon = 0.2$,
$\gamma = 0.97$, $\alpha = 0.5$, and an LSTM hidden size of 256. Deviations from these
values result in either instability (large $\epsilon$), myopic decision-making (low $\gamma$),
or suboptimal trade-offs between cost and reliability (extreme $\alpha$ values). 
The relatively smooth performance variations around the optimal points indicate that the
proposed framework is robust to moderate hyperparameter perturbations.

\section{Conclusion}

This paper presented a comprehensive Deep Reinforcement Learning-based framework for
green energy management in e-commerce data centers. By formulating the energy dispatch
problem as a Markov Decision Process and training a Proximal Policy Optimization agent
augmented with a hybrid LSTM-Attention architecture, the proposed system learns adaptive
control policies that dynamically balance renewable energy sources, battery storage, and
grid electricity in response to continuously evolving workload demands and generation
patterns.

Experimental evaluations conducted on three real-world datasets demonstrated the
superiority of the proposed approach over six representative baselines spanning
rule-based heuristics, classical reinforcement learning, and state-of-the-art deep
reinforcement learning methods. The proposed framework achieves a 38\% reduction in
energy costs relative to conventional heuristic strategies and surpasses the strongest
DRL baseline, GreenDRL, by 4.6\% in cost efficiency. Simultaneously, it maintains an
SLA violation rate of only 1.5\%, an energy efficiency of 83.7\%, and reduces carbon
emissions by over 40\% compared to rule-based approaches. These performance gains are
consistently reflected across all six evaluation metrics, as evidenced by the
multi-criteria radar analysis presented in this work.

The ablation study confirmed that each component of the proposed architecture makes a
meaningful contribution to overall performance. In particular, the LSTM-based temporal
modeling module proved critical for capturing sequential dependencies in workload and
renewable generation data, while the temporal attention mechanism further improved the
agent's ability to selectively focus on the most informative historical observations.
The multi-objective reward formulation was shown to be essential for simultaneously
optimizing cost efficiency, environmental impact, and service reliability. The
hyperparameter sensitivity analysis further demonstrated that the proposed framework is
robust to moderate variations in its configuration, lending confidence to its
applicability in real operational environments.

Despite these encouraging results, several limitations warrant consideration. The
current framework relies on simulated environments constructed from historical data,
and its direct deployment in live data center infrastructures would require careful
integration with real-time monitoring systems and safety-critical control layers.
Furthermore, the scalability of the approach to geographically distributed data center
networks involving multi-site coordination and heterogeneous grid pricing structures
remains an open challenge.

Future research directions include the extension of the proposed framework to
multi-agent settings, where geographically distributed data centers can collaboratively
optimize their collective energy consumption while accounting for regional renewable
availability and dynamic electricity markets. Incorporating probabilistic forecasting
models for renewable generation and workload prediction as auxiliary inputs to the DRL
agent represents another promising avenue for further improving decision quality under
uncertainty. Finally, exploring transfer learning strategies to accelerate policy
adaptation across data center configurations with differing hardware profiles and
energy mixes would enhance the practical deployability of the framework. These
extensions would contribute to advancing the broader goal of achieving carbon-neutral,
intelligent, and resilient digital infrastructures.

\section*{Ethical Approval} 
Not Applicable.

\section*{Conflict of Interest}
The authors declare that they have no known competing financial interests or personal relationships that could have appeared to influence the work reported in this paper.

\section*{Funding}
This research received no external funding.

\section*{Data Availability}
Data will be made available on request.

\end{document}